\documentclass[11pt,a4paper]{article}
\usepackage[hyperref]{ranlp2023}
\usepackage{times}
\usepackage{latexsym}
\usepackage{graphicx}

\usepackage{microtype}

\aclfinalcopy 

\title{Classification of US Supreme Court Cases using BERT-Based Techniques}

\author{Shubham Vatsal \\
  New York University, CIMS \\
  New York, USA \\
  \texttt{sv2128@nyu.edu} \\ \And
  Adam Meyers \\
  New York University, CIMS \\
  New York, USA \\
  \texttt{meyers@cs.nyu.edu} \\ \AND
  John E. Ortega \\
  Northeastern University \\
  Boston, USA \\
  \texttt{j.ortega@northeastern.edu} \\
  }

\date{}

\begin{document}
\maketitle
\begin{abstract}
Models based on bidirectional encoder representations from transformers (BERT) produce state of the art (SOTA) results on many natural language processing (NLP) tasks such as named entity recognition (NER), part-of-speech (POS) tagging etc. An interesting phenomenon occurs when classifying long documents such as those from the US supreme court where BERT-based models can be considered difficult to use on a first-pass or out-of-the-box basis. In this paper, we experiment with several BERT-based classification techniques for US supreme court decisions or supreme court database (SCDB) and compare them with the previous SOTA results. We then compare our results specifically with SOTA models for long documents. We compare our results for two classification tasks: (1) a broad classification task with 15 categories and (2) a fine-grained classification task with 279 categories. Our best result produces an accuracy of 80\% on the 15 broad categories and 60\% on the fine-grained 279 categories which marks an improvement of 8\% and 28\% respectively from previously reported SOTA results. 
\end{abstract}

\section{Introduction}
Every October, the US supreme court begins a new term on the first Monday. Each term includes a number of significant and complex cases that address a variety of issues, including environmental protection law, free speech, equal opportunity, tax law etc. Legal court decisions are lawful statements acknowledged by a judge providing the justification and reasoning for a court ruling. The court's decisions not only have repercussions on the people involved in the case but also on the society as a whole. During a regularly scheduled court session, the justice who wrote the main decision sums it up from the bench. A copy of the opinion is then quickly posted on the concerned website. Legal experts must search through and classify these court decisions to support their research. This can require a lot of manual effort. Artificial intelligence (AI) and machine learning (ML) can significantly reduce the burden of this type of manual work. Washington University's collection of 8419 manually labeled supreme court documents (SCDB)  \cite{spaeth2013supreme} provides the basis for bench marking this type of AI task.

There are multiple challenges involved when dealing with SCDB or legal documents of similar kinds. First, these documents are exceptionally long which causes numerous difficulties. For example, feature vectors can be too large to fit in the memory or contextual variance can be too high for models to handle. Next, the decision of a case requires identification of other relevant cases that can support the decision of the current case, which usually involve similar circumstances. Therefore, it is of utmost importance to identify the similarities in terms of legal aspects while classifying the cases in the same category. Finally, understanding of these legal documents requires supervision of legal expert. Legal statements, like other specialized domains, have unique characteristics when compared to other generic corpora, such as unique vocabulary, exclusive syntax, idiosyncratic semantics etc.
Vocabulary, syntax, semantics and other linguistic characteristics can be specific to the legal domain or even the subdomain of court decisions, thus necessitating tools that are trained specifically for such domains. 
Hence, an automatic system to categorize and process these documents is extremely useful.

In this paper, we explore many novel techniques to counter the problem faced by models similar to BERT \cite{devlin2018bert} when dealing with documents of length more than 512 tokens. Some of the techniques include: analysing which 512 token chunks of documents make the best contribution towards classification, using summarized version of documents for classification, a voting-based ensemble approach, classification based on concatenation of 512 token chunks. We compare our results with previous non-BERT like models as well as models which can take inputs of length more than 512 tokens.

The rest of the paper is organized in the following way. Section 2 talks about related work. Section 3 talks about the SCDB. We describe the working of BERT-based techniques in Section 4. Section 5 concentrates on the experiments we conducted and the corresponding results we achieved. The final section discusses future work.
\section{Related Work}

Recently, there has been a surge in research in the domain of NLP associated with the legal documents. \cite{dragoni2016combining} talks about combining linguistic information from WordNet with a syntax-based retrieval of rules from legal text along with logic-based retrieval of dependencies from chunks of such texts leading to extraction of machine-readable rules from legal documents. 
\cite{zhong2020does} illustrates several embedding-based and symbol-based approaches for judgement prediction, legal question answering and similar case matching. \cite{dale2019law} discusses five areas of legal activity where NLP is playing an important role. These areas include legal research for finding information relevant to a legal decision, electronic discovery determining the relevance of documents in an information request, contract review to check that a contract is complete, document automation to generate routine legal documents and legal advice using question-answering dialogues. \cite{kanapala2019text} discusses different available approaches for summarization of legal texts and compares their performances on various datasets. \cite{garcia2017cliel} showcases a scalable and flexible information extraction method, aimed at extraction of information from legal documents regardless of format, layout or structure, by considering the context.
\cite{yeung2019effects} comes up with a German legal BERT model and evaluates its performance on downstream NLP tasks including classification, regression and similarity.

BERT-based models have shown some ground breaking performance in many NLP tasks. Nowadays, they are being widely used in the legal domain as well. \cite{shao2020bert} proposes BERT-PLI to capture the semantic relationships at the paragraph-level and then goes on to infer the relevance between two legal cases by aggregating paragraph-level interactions. \cite{chalkidis2020legal} releases Legal-BERT, a family of BERT models for the legal domain intended to assist legal NLP research. \cite{sanchez2020easing} studies a case in the context of legal professional search and presents how BERT-based approach outperforms other traditional approaches. \cite{chau2020vnlawbert} proposes an answer selection approach by fine-tuning BERT on their Vietnamese legal question-answer pair corpus. They further pre-train BERT on a Vietnamese legal domain-specific corpus and show that this new BERT performs better than the fine-tuned BERT. 

Classification of legal documents is an important NLP task which can automate alignment of legal documents with human-defined categories. \cite{elwany2019bert} classifies a proprietary corpus consisting of hundreds of thousands of legal agreements using BERT. \cite{limsopatham2021effectively} compares multi-label and binary classification of legal documents using variances of pre-trained BERT-based models and other approaches to handle long documents. Our work falls somewhat along similar lines but we use a different dataset of legal documents with considerably different properties. Moreover, many of our BERT-based techniques are significantly different from \cite{limsopatham2021effectively}. \cite{de2020victor} presents baseline results for document type classification and theme assignment, a multi-label problem using their newly built Brazil’s supreme court digitalized legal documents dataset. \cite{li2019combining} proposes a method for learning Chinese legal document classification using graph long short-term memory (LSTM) combined with domain knowledge extraction. 
\cite{vsaric2014multi} addresses multi-label classification of Croatian legal documents using EuroVoc thesaurus.
\cite{howe2019legal} experimented classification of Singapore supreme court judgments using topic models, word embedding feature models and pre-trained language models. \cite{mumcuouglu2021natural} presents results on predicting the rulings of the Turkish Constitutional Court and Courts of Appeal using fact descriptions. 
\cite{sulea2017predicting} investigates various text classification techniques to predict French Supreme Court decisions whereas \cite{virtucio2018predicting} does similar work for Philippine Supreme Court. 

SCDB has been used in various prominent NLP tasks. \cite{silveira2021topic} uses supreme court data and performs topic modelling using domain-specific embeddings. These embeddings are obtained from pre-trained Legal-BERT. \cite{katz2017general} constructs a model to predict the voting behavior of US supreme court and its justices in a generalized, out-of-sample context by using SCDB along with some other derived features. \cite{chalkidis2021lexglue} talks about classification performance of different BERT-based as well as non-neural architectures on many datasets of legal domain including SCDB. Our experiments differ from their work in multiple ways. First, they don't analyze the performance of these models across fine-grained 279 categories which is a harder classification task. Second, they do not experiment different techniques to tackle the problem of restricted input sequence length of 512 tokens in BERT-based models which is one of the key points of our work. The analysis of these techniques helps us in achieving SOTA across both broad as well as fine-grained classification tasks. Finally, the version of SCDB used by them differs from what we have used in our experiments. There is a significant overlap but the versions are not exactly the same.
\cite{undavia2018comparative} presents classification of SCDB across broad 15 categories and fine-grained 279 categories. Our work exactly aligns with this work but the usage of BERT-based techniques helps us in achieving better results. 

\section{Data}

Our paper is primarily based on classification of US supreme court decisions dataset or supreme court database from Washington University School of Law \cite{spaeth2013supreme}. Documents are classified by topic in a 2 level ontology, providing the basis for two different classification tasks: one using 15 broad category labels and another using 279 fine-grained category labels. The general statistics associated with this dataset can be found in Table \ref{datastats}. As we can see from Fig. 1 and Fig. 2 of \cite{undavia2018comparative}, SCDB is highly imbalanced in terms of number of data points per label in both classification tasks. Before we conduct our experiments, we apply a pre-processing step to remove footnotes.\footnote{We provisionally assume that footnotes constitute noise.} 
The SCDB dataset poses some difficulties for BERT-based classification task because
the average length of SCDB documents is much longer than most of the other legal datasets used for classification task. For example, European Court of Human Rights (ECHR) \cite{chalkidis2021paragraph} and Overruling \cite{zheng2021does} datasets have only 1662.08 and 21.94 mean length in comparison to SCDB's mean length of 6960.60 tokens.

\begin{table}
\centering
\caption{\label{datastats} Data Statistics }
\begin{tabular}{cc}
\hline \textbf{Metric} & \textbf{Value} \\ \hline
Dataset Size & 8419 \\
Min \# Tokens & 0\\
Max \# Tokens & 87246\\
Median \# Tokens & 5552\\
Mean \# Tokens & 6960.60\\
Min \# Tokens After Pre-Processing & 0\\
Max \# Tokens After Pre-Processing & 87246\\
Median \# Tokens After Pre-Processing & 3420 \\
Mean \# Tokens After Pre-Processing & 4458.15\\
\hline
\end{tabular}

\end{table}

\begin{table*}
\centering
\caption{Results For Best-512 \label{tab:resultschunk}}
\begin{tabular}{ ccccc}
\hline
\bfseries {Chunk I} & \bfseries{Labels} & \bfseries{Accuracy}  & \bfseries{Precision} & \bfseries{F1}  \\

\hline
  & 15 &  \bf{0.747}  & \bf{0.752}  & \bf{0.744} \\
Chunk 1 & 279 &  \bf{0.545}  & \bf{0.498}  & \bf{0.500} \\
\hline
  & 15  & 0.688  & 0.692   & 0.683\\
Chunk 2 & 279  & 0.464  & 0.417   & 0.419\\
\hline
  & 15 & 0.683   & 0.682  & 0.673\\
Chunk 3 & 279 &  0.457  & 0.408  & 0.409\\
\hline

  & 15 & 0.704   & 0.702  & 0.696\\
Chunk 4 & 279 &  0.459  & 0.405  & 0.412\\
\hline

  & 15 & 0.687   & 0.685  & 0.682\\
Chunk 5 & 279 &  0.452  & 0.404  & 0.406\\
\hline

  & 15 & 0.679   & 0.689  & 0.676\\
Chunk 6 & 279 &  0.454  & 0.411  & 0.409\\
\hline

\end{tabular}
\end{table*}

\begin{table*}
\centering
\caption{Results For Stride-64, 128 \label{tab:resultsstride}}
\begin{tabular}{ ccccc }
\hline
\bfseries {Stride} & \bfseries{Labels} & \bfseries{Accuracy}  & \bfseries{Precison}  & \bfseries{F1}  \\

\hline
  & 15 &  \bf{0.774}  & 0.777  & \bf{0.771} \\
64 & 279 &  \bf{0.563}  & \bf{0.514}  & \bf{0.519} \\
\hline
  & 15  & 0.762  & \bf{0.779}  & 0.763\\
128 & 279  & 0.557  & 0.505  & 0.510\\
\hline

\end{tabular}
\end{table*}

\section{Proposed Techniques}

We explore various BERT-based techniques to classify SCDB in this section. We apply these individual techniques to both the classification tasks i.e one with 15 categories and the other one with 279 categories. We use either BERT or different versions of BERT for our classification tasks. Particularly, we use BERT \cite{devlin2018bert},  RoBERTa \cite{liu2019roberta} and Legal-BERT \cite{chalkidis-etal-2020-legal} which have the restriction of maximum input sequence length of 512 tokens. We first try out our different techniques using BERT, RoBERTa and Legal-BERT. Later, we compare the results of these techniques with other transformer-based models like LongFormer \cite{beltagy2020longformer} and Legal-Longformer \footnote{\url{https://huggingface.co/saibo/legal-longformer-base-4096}} which accept longer sequences.














\subsection{Best-512}
Let $c_{i}$ represent the $i^{th}$ 512-length chunk of a given document in SCDB. We have taken the length of each chunk to be 512 because that is the maximum length of a sequence accepted by a BERT-based model. We calculate our evaluation metrics on these chunks and thus analyse which chunk of documents contribute the most towards their correct classification. An important point to consider here is that the evaluation metrics are calculated on the best averaged chunk and not on different best chunks for individual data points. Since the median length of documents in SCDB is around 3000, we compute the performance of our three BERT-based models on $c_{1}, c_{2}, c_{3}, c_{4}, c_{5}$ and $c_{6}$ where $c_{1}$ represents $1^{st}$ 512 length chunk of documents, $c_{2}$ represents the $2^{nd}$ 512 length chunk of documents and so on. When the length of a document is less than i*512, we take the last 512 or less than 512 (when i=1) tokens of the document. Initially, we use BERT to find the best averaged chunk $c_{i}$ for all the documents and later use the same $c_{i}$ to experiment with different BERT-based models discussed in Section \ref{expr}. The result showing the best averaged chunk using BERT can be seen in Table \ref{tab:resultschunk}. The final result comparing the performance of different BERT-based models using the best averaged chunk obtained from Table \ref{tab:resultschunk} for both the classification tasks can be seen in Table \ref{results15labels} and Table \ref{results279lables}.

\subsection{Summarization-512}
In this technique, we summarize the documents of SCDB in 512 tokens. We use the summarization pipeline from Hugging Face with default parameters. The maximum sequence length that this summarization model can accept is 1024. So, we first convert a document to some splits based on the length of that document. The number of splits $n_i$ for a given document $d_i$ is defined as $l_i$/1024 where $l_i$ is the length of the document. Now, since the total length of the summarized version of a document can only be upto 512 tokens long, we further calculate the number of tokens per split $nw_i$ for all the splits of a given document $d_i$. The number of tokens per split $nw_i$ is calculated as 512/$nw_i$. Finally, we concatenate all $nw_i$'s of a given document $d_i$ to get the final summarized version. Let's go through an example to make it more clear. Let's say we have document $d_i$ of length $l_i$ 4096. For this document, the number of splits $n_i$ is going to be 4 whereas the number of tokens per split is going to be 128. So, we summarize each split into 128 tokens and finally concatenate all 4 summarized versions (128*4) to create a final summarized version of 512 tokens . These summarized versions are then used for both the classification tasks. The results are shown in Table \ref{results15labels} and Table \ref{results279lables}.
\begin{table*}
\centering
\caption{Results For 15 Categories \label{results15labels}}
\begin{tabular}{ ccccc }
\hline
\bfseries {Technique} & \bfseries{Model} & \bfseries{Accuracy}  & \bfseries{Precision} & \bfseries{F1}  \\
\hline
  & BERT  &  0.747  & 0.752  & 0.744 \\
Best-512 & RoBERTa &  0.768  & 0.773  & 0.766 \\
  & Legal-BERT &  0.785  & 0.795  & 0.785\\
\hline
  & BERT  & 0.736  & 0.748   & 0.734\\
Summarization-512 & RoBERTa  & 0.745  & 0.761   & 0.747\\
  & Legal-BERT &  0.789  & 0.801  & 0.790\\
\hline
  & BERT & 0.772  & 0.775  & 0.769\\
Concat-512 & RoBERTa &  0.772  & 0.782  & 0.773\\
  & Legal-BERT & 0.791 & 0.799  & 0.791\\
\hline

  & BERT & 0.755   & 0.755  & 0.752\\
Ensemble & RoBERTa &  0.766  & 0.770 & 0.763\\
  & Legal-BERT & 0.782 & 0.792  & 0.782\\
\hline

  & BERT & 0.774  & 0.777  & 0.771\\
Stride-64 & RoBERTa &  0.779  & 0.785  & 0.778\\
  & Legal-BERT & \bf{0.801} & \bf{0.805}  & \bf{0.800}\\
\hline

LSMs & LongFormer &  0.742  & 0.753  & 0.739\\
  & Legal-LongFormer & 0.775 & 0.785  & 0.775\\
  & CNN \cite{undavia2018comparative} & 0.724 & -  & -\\
\hline

\end{tabular}

\end{table*}
\subsection{Concat-512}
Let $c_{i}$ represent the $i^{th}$ 512-length chunk of a given document in SCDB. In this technique, we accept $i$ parallel inputs of 512 sequence length. Corresponding to $i$ parallel inputs we have $i$ BERT-based models which are trained simultaneously and their outputs are concatenated. In a sense, we concatenate i CLS tokens from i BERT-based models.  This concatenated output is then fed into a dense layer with softmax activation and number of units being equal to 15 or 279 based on the classification task. Again, because the median length of documents is around 3000, for this experiment we only take $c_{1}, c_{2}, c_{3}, c_{4}, c_{5}$ and $c_{6}$ into consideration where $c_{1}$ represents $1^{st}$ 512 length chunk of documents, $c_{2}$ represents the $2^{nd}$ 512 length chunk of documents and so on. One important question that needs to answered here is what happens when the length of a document is less than 3000 tokens. Let's go through an example to understand this. Let's say we have document $d_i$ of length $l_i$ 1024. In this case, only the first 2 BERT-based models actually receive a valid input whereas the other 4 BERT-based models receive a null input. So, in cases where the length of the document is less than 3000, one could point out that what if the dense layer ends up learning the prediction of labels based on just presence and absence of last few chunks of the documents. Our justification for this point is that this could happen only for a very small number of documents as the median length of SCDB is 3000 and hence this scenario will not affect the generalization capabilities of the model.
The results can be seen in Table \ref{results15labels} and Table \ref{results279lables}.







\subsection{Ensemble}
Let $c_{i}$ represent the $i^{th}$ 512-length chunk of a given document in SCDB. In our ensemble approach, we train a model $m_i$ for each $c_i$. During testing, we predict the final label of a document using a maximum voting mechanism where the final prediction is what the majority of $m_i$ end up choosing which is given by equation \ref{eq1}. The ${{m_i}_{,t}}$ term in equation \ref{eq1} can take either the value of 0 or 1 based on its prediction.  If $i^{th}$ classifier chooses class t, then ${{m_i}_{,t}}$ = 1, and 0, otherwise. The $\#nc$ term in equation \ref{eq1} refers to the number of classes for the corresponding classification task.  In the case where the length of a document is less than i*512, we ignore that document for the training of that $m_i$. Since the median length of documents in SCDB is around 3000, for this experiment we only take $c_{1}, c_{2}, c_{3}, c_{4}, c_{5}$ and $c_{6}$ into consideration where $c_{1}$ represents $1^{st}$ 512 length chunk of documents, $c_{2}$ represents the $2^{nd}$ 512 length chunk of documents and so on. As stated previously, we run this experiment on different BERT-based models and note down the results for both the classification tasks. The results can be seen in Table \ref{results15labels} and Table \ref{results279lables}.

\begin{equation} \label{eq1}
label\ =\  argmax_{t \in \{1,2..\#nc\}} \Sigma ^{6}_{i=1} \ m_{i,t}
\end{equation}

\subsection{Stride-64, 128}

Let $c_{ij}$ represent the $i^{th}$ 512-length chunk of a given document $d_j$ in SCDB. Stride technique takes into consideration a window of tokens which is shared amongst any two consecutive chunks $c_{ij}$ and $c_{ij+1}$ contrary to what is observed in Ensemble and Concat-512 techniques where there is a contextual boundary between any two consecutive chunks. Let's take an example of Stride-64 where the length of shared window of tokens is 64 to develop more clarity. Let $c_{ij}[0:512]$ represent the 512 tokens present in $c_{ij}$. Following the idea of Stride technique, for the first two chunks  $c_{1j}[0:512]$ and $c_{2j}[512:1024]$, $c_{1j}[448:512]$ and $c_{2j}[0:64]$ tokens of $c_{1j}$ and $c_{2j}$ respectively are going to be exactly same as the length of shared window of tokens is 64. So, if we have a document $d_j$ of length 1024 tokens, we will have three $c_{ij}'s$ with $c_{1j}[448:512]=c_{2j}[0:64]$ and $c_{2j}[448:512]=c_{3j}[0:64]$. Also, to elaborate $d_j[0:512]=c_{1j}[0:512]$, $d_j[448:512]=c_{2j}[0:64]$, $d_i[512:960]=c_{2j}[64:512]$, $d_j[896:960]=c_{3j}[0:64]$ and $d_j[960:1024]=c_{3j}[64:128]$. We have pad tokens in 
$c_{3j}[128:512]$. Taking the median length of documents of SCDB into consideration, we again experiment up till length around 3000 tokens as explained for other techniques previously. Initially, we use BERT model to find the best shared window size for all the documents and later use the same shared window size to experiment with different BERT-based models discussed in Section \ref{expr}. The result showing the best shared window size using BERT can be seen in Table \ref{tab:resultsstride}. The final result comparing the performance of different BERT-based models using the best shared window size obtained from Table \ref{tab:resultsstride} for both the classification tasks can be seen in Table \ref{results15labels} and Table \ref{results279lables}.

\begin{table*}
\centering
\caption{Results For 279 Categories \label{results279lables}}
\begin{tabular}{ ccccc }
\hline
\bfseries {Technique} & \bfseries{Model} & \bfseries{Accuracy}  & \bfseries{Precision}  & \bfseries{F1}  \\
\hline
  & BERT &  0.545  & 0.498  & 0.500 \\
Best-512 & RoBERTa &  0.533  & 0.474  & 0.480 \\
  & Legal-BERT &  0.586  & 0.554  & 0.547\\
\hline
  & BERT  & 0.529  & 0.486   & 0.483\\
Summarization-512 & RoBERTa  & 0.522  & 0.456   & 0.466\\
  & Legal-BERT &  0.585  & 0.553  & 0.549\\
\hline
  & BERT & 0.554   & 0.511  & 0.511\\
Concat-512 & RoBERTa &  0.534  & 0.460  & 0.475\\
  & Legal-BERT & 0.596 & 0.560 & 0.559\\
\hline

  & BERT & 0.520   & 0.464  & 0.471\\
Ensemble & RoBERTa &  0.520  & 0.455  & 0.467\\
  & Legal-BERT & 0.553 & 0.529  & 0.520\\
\hline

  & BERT & 0.563  & 0.514  & 0.519\\
Stride-64 & RoBERTa &  0.536  & 0.465  & 0.479\\
  & Legal-BERT & \bf{0.609} & \bf{0.584}  & \bf{0.575}\\
\hline

LSMs & LongFormer &  0.534  & 0.481  & 0.487\\
  & Legal-LongFormer & 0.562 & 0.515  & 0.519\\
  & CNN \cite{undavia2018comparative} & 0.319 & -  & -\\
\hline

\end{tabular}

\end{table*}

\subsection{Longer Sequence Model (LSM)}

These are the models which can accept input sequence longer than 512 tokens. We ran our experiments with two such models, LongFormer and Legal-Longformer. Apart from these two models, we also use the results reported by \cite{undavia2018comparative} where the best performing model uses convolutional neural network (CNN) architecture.







\section{Experiments \& Results}
\label{expr}
\begin{figure}[h]
    \centering
    \includegraphics[scale=0.29]{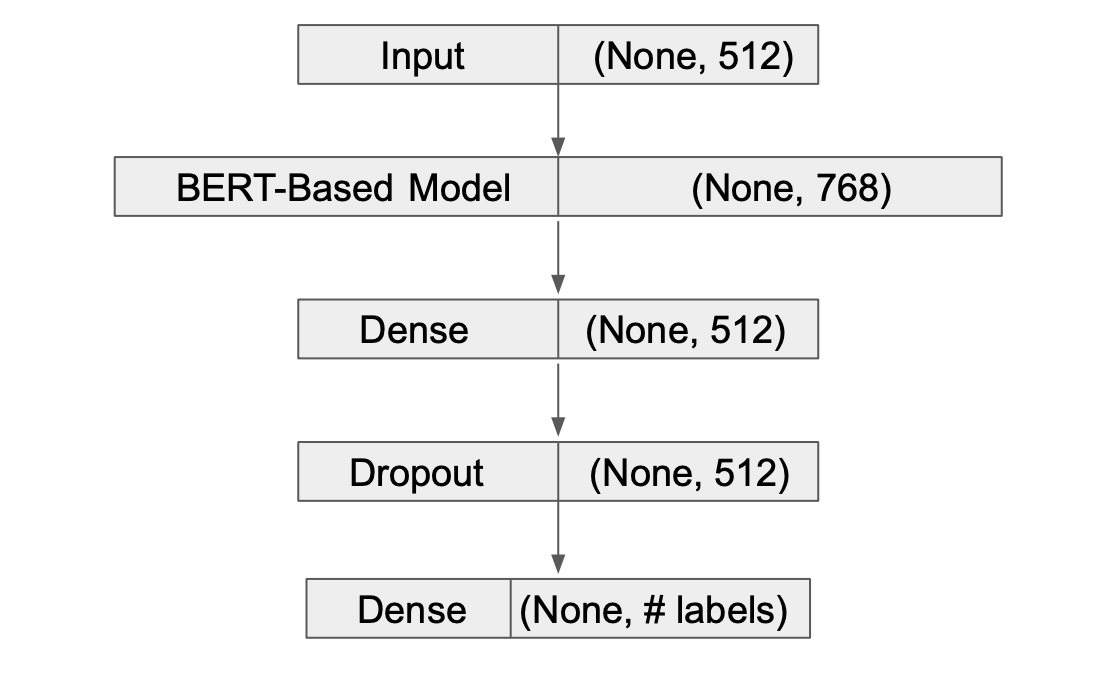}
    \caption{Best-512, Summarization-512, Ensemble, LSMs General Architecture}
    \label{fig:general}
\end{figure}

\begin{figure}[h]
    \centering
    \includegraphics[scale=0.25]{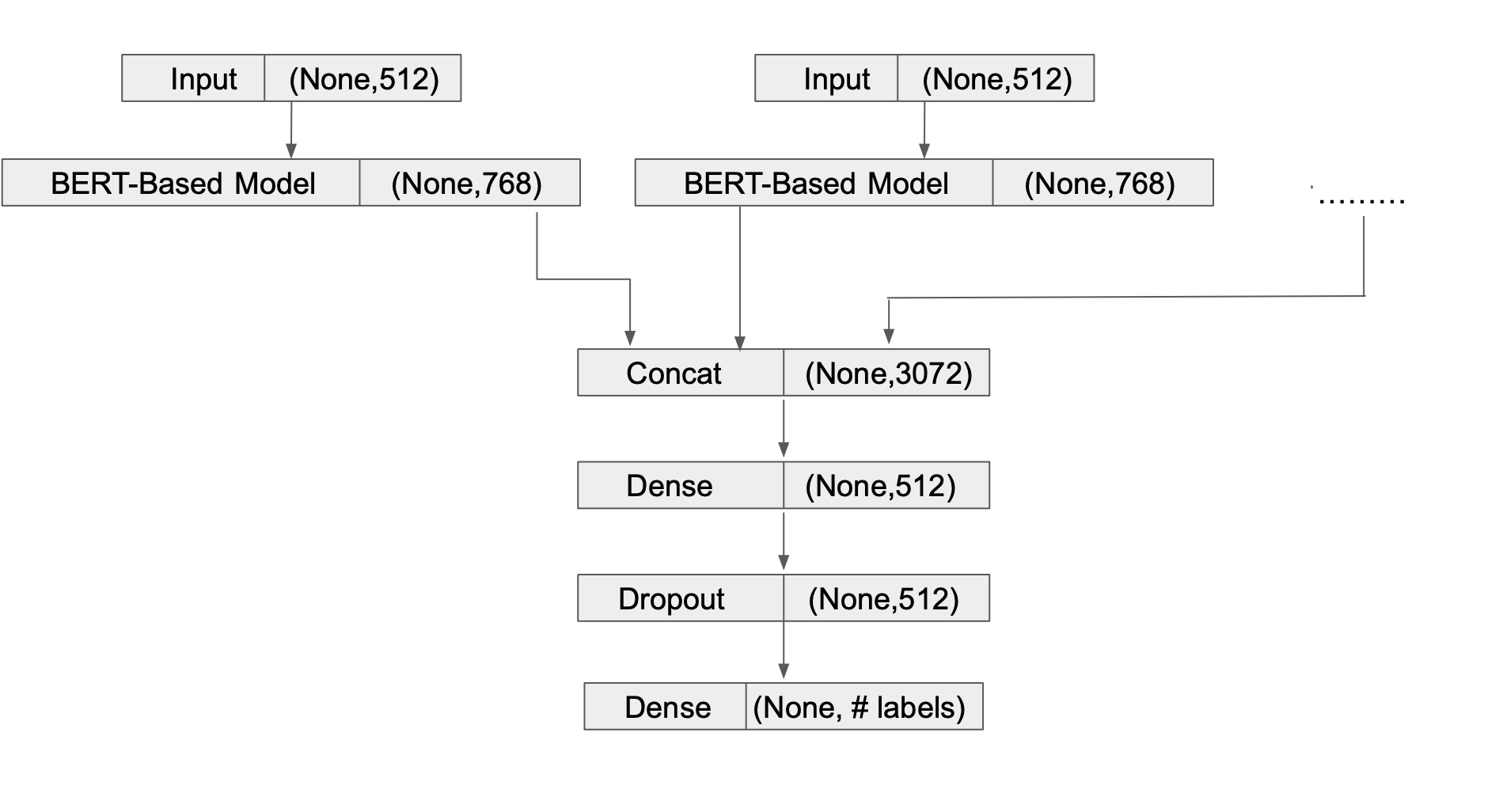}
    \caption{Concat-512, Stride-64 General Architecture}
    \label{fig:concat}
\end{figure}
The code 
\footnote{\href{https://github.com/shubham30vatsal/Web-Of-Law}{Web-Of-Law-Experiments}} related to all the experiments discussed below have been made public. We use weighted F1, accuracy and weighted precision as our evaluation metrics for both the classification tasks. The hyper-parameters used for all the above techniques except RoBERTa include batch size to be 8, number of epochs as 5, learning rate to be 3e-5 and loss to be Categorical Cross Entropy. For RoBERTa, we keep all the hyper-parameters to be the same except the learning rate which is changed to 1e-5. We split 8419 data points in 90:10 ratio of train and test sets. We run each experiment 5 times and take the average of the best epoch score to get the final score. We choose Adam \cite{kingma2014adam} as our optimizer. We use bert-base-uncased version of BERT, roberta-base version of RoBERTa, legal-bert-base-uncased version of Legal-BERT, longformer-base-4096 version of Longformer and legal-longformer-base-4096 version of Legal-Longformer from Hugging Face \footnote{\url{https://github.com/huggingface/transformers}}. All the BERT-based models accept a sequence of maximum of 512 tokens whereas Longformer and Legal-Longformer accept a sequence of maximum of 4096 tokens. 
Figure \ref{fig:general} shows the general architecture of Best-512, Summarization-512, Ensemble and LSMs with some differences in corresponding dimensions and input type. Similarly, the general architecture of Concat-512 and Stride-64 with some differences in corresponding dimensions and input type can be seen in Figure \ref{fig:concat}. Each rectangular box in the image is divided into two parts. The left part of the box represents the name of the layer whereas the right part shows the output dimension of the layer. The architecture of both the types of models is mostly similar except for the form in which they accept their inputs.

As we can see from Table \ref{results15labels} and \ref{results279lables}, a BERT-based model which has been trained on legal data like Legal-BERT or other transformer-based model with it's training data coming from legal domain like Legal-Longformer always outperforms other models within a given technique. When comparing the best performing model across different BERT-based techniques for 15 categories, the techniques can be ranked as Stride-64 giving the best result, followed by Concat-512, followed by Summarization-512, followed by Best-512, followed by Ensemble and finally we have LSMs. Similarly, when comparing the best performing model across different BERT-based techniques for 279 categories, the techniques can be ranked as Stride-64 giving the best result, followed by Concat-512, followed by Summarization-512, followed by Best-512, followed by Ensemble and finally we have LSMs. There could be multiple reasons for LSMs to not have a better performance than other models. One, LSMs are designed in a way to allow multi-head attentions to adhere to a restricted window contrary to BERT-based models where these multi-head attentions are free to concentrate on any of the tokens. Second, with more number of tokens, more variance is created which can lead to poor performance. We have already seen from Table \ref{tab:resultschunk}, it is just the first 512 tokens which contribute the most towards the classification of the corresponding documents. The rationale behind the first 512 tokens to contribute the most towards classification can be attributed to the fact that the documents are of unequal length and there are many documents which are less than or equal to the length of 512 tokens. Also, the reason why Best-512 performs poorly is because even though it is the first 512 token chunk which contributes the most for these classification tasks but still the context beyond these 512 tokens does make an impact. Summarization-512 tries to capture the context across the entire document but due to its limitation to express this context in just 512 tokens, it is not as efficient as Concat-512, Stride-64 or Ensemble. Ensemble outperforms Best-512 and Summarization-512 because it tries to exploit joint learning across multiple 512 token chunks through its maximum voting mechanism rule. Concat-512 on the other hand captures better context across multiple 512 token chunks as it learns this knowledge during back propagation of the model. Finally, Stride-64 outperforms Concat-512 because when we take disjoint chunks, the continuity of context goes missing whereas if there is an overlapping portion of text between two consecutive chunks, it gives better contextual understanding.


\section{Conclusion \& Future Work}

In this paper, we experimented with various BERT-based techniques on SCDB and presented the corresponding results comparing them with other state of the art models. We further did an analysis on how even with the given restriction of the input sequence length of 512 tokens for BERT-based models, we can leverage these techniques to get some improvement.

As a part of future work, we can leverage the knowledge embedded in references of a given SCDB document. A reference in an SCDB document basically refers to some other SCDB document that has been cited to legally justify the decision taken on the former SCDB document. The raw text of these references may not be very helpful in improving the classification tasks. We can use a graph structure to denote the relations between a given SCDB document with other documents cited in it. The final classification result of a given SCDB document can be calculated based on some form of aggregation incorporating classification results of its references weighted by the graph structure representing quantified relations with the SCDB document at hand.

Another area of future work can be applying greedy approaches to the techniques discussed in this work. For example, we can have a greedy summarization technique where in the final 512 token summary, we can include more number of tokens from the best performing 512 token chunk as inferred from Best-512 technique. Similarly, we can have greedy ensemble technique where during the voting phase, we can give more weight to the best performing 512 token chunk as the results show from Best-512 technique.

\bibliographystyle{acl_natbib}
\bibliography{ranlp2023}

\end{document}